\documentclass{article}
\usepackage{ijcai26}
\usepackage{times}
\usepackage{url}
\usepackage[hidelinks]{hyperref}
\usepackage[utf8]{inputenc}
\usepackage[small]{caption}
\usepackage{graphicx}
\usepackage{amsmath}
\usepackage{booktabs}
\title{Efficiency Measured in English: Tokeniser Coverage Bounds\\ Small Language Models on Indic Scripts}

\author{
Anna Pramod
\affiliations
School of Computer Science, University of Leeds, United Kingdom
\emails
rthk0461@leeds.ac.uk
}

\begin{document}
\maketitle

\begin{abstract}
Small Language Models with quantisation and parameter-efficient adaptation are widely offered as the route to multilingual AI under compute constraints, and their efficiency is almost always reported as a single memory figure. We show that this figure is measured in English and does not transfer. Across a 131k-sentence corpus spanning eleven Indic scripts, TinyLlama-1.1B's~\cite{zhang2024tinyllama} tokeniser emits 1.34 subword tokens per Latin-script word and \textbf{22.60 per Kannada word}. This is not inefficiency; it is absence. Its vocabulary contains \textbf{0\% of Odia's codepoints}, 1\% of Telugu's and 2\% of Kannada's, and for five of eleven scripts the measured fertility sits at the UTF-8 byte-fallback ceiling, on Odia at \textbf{99.6\% of it}, with \textbf{93.6\% of emitted tokens being raw bytes}. Because SentencePiece byte fallback never raises an unknown-token marker, \textbf{this failure is entirely silent}: TinyLlama's UNK rate is 0.0\% on every script, including the one it reads as bytes. We further show that low fertility can indicate the \emph{opposite} of efficiency: mBERT posts its best score anywhere (1.10 tokens/word, on Odia) while discarding 87.9\% of the text as UNK. MuRIL, an Indic-vocabulary encoder with a fifth of TinyLlama's parameters, reads every script at 1.23--2.26 tokens/word with zero UNK and zero byte fallback. \textbf{The binding constraint is the vocabulary, not the parameter count.} The consequence is concrete: on a 2048-token context window, the same model gives an English user 1{,}528 words of usable context and a Kannada user \textbf{91}.
\end{abstract}

\section{Introduction}

That low-resource languages are underserved~\cite{joshi2020state}, and that frontier-model compute is out of reach for the institutions that would most benefit~\cite{strubell2019energy,schwartz2020green}, are premises this workshop does not need re-argued. We take them as given and ask what follows.

The standard answer is a small, quantised, parameter-efficiently adapted model, sold on a single number: memory saved. We argue that number is close to meaningless as a deployment signal, for a reason that is specific, measurable in minutes on a laptop, and cheap to fix.

\textbf{The efficiency of a multilingual SLM is measured in English, and it does not transfer.} TinyLlama~\cite{zhang2024tinyllama} reads an English word in 1.34 subword tokens and a Kannada word in 22.60. Whatever is saved on weights is spent seventeen times over on sequence length, in precisely the languages the system exists to serve. The saving and the cost are denominated in different units, and only one of them is ever published.

Our contributions:
\begin{itemize}
    \item \textbf{The vocabulary is not inefficient, it is empty} (\S\ref{sec:vocab}). For five of eleven Indic scripts, TinyLlama's measured fertility sits at the UTF-8 byte-fallback ceiling. Its vocabulary contains 0--2\% of the codepoints of Odia, Telugu, Kannada and Gujarati. It is not tokenising these scripts; it is consuming raw bytes.
    \item \textbf{The failure is silent} (\S\ref{sec:silent}). Byte fallback emits \emph{zero} UNK tokens on every script we tested, including the one it reads as 94\% raw bytes. No error, no warning, no signal in any standard metric.
    \item \textbf{Low fertility can mean total data destruction} (\S\ref{sec:unk}). mBERT's best fertility score in our study is 87.9\% UNK. Fertility must never be reported without an UNK/byte rate beside it.
    \item \textbf{The cost, in the user's terms} (\S\ref{sec:context}). A 2048-token window is 1{,}528 words of English and 91 words of Kannada. This is disclosed nowhere.
    \item \textbf{An honest corpus characterisation} (\S\ref{sec:setup}), correcting two claims widely made about a popular Indic dataset.
\end{itemize}

\section{Setup, and What the Corpus Actually Is}
\label{sec:setup}

We analyse a widely-used 131{,}044-sentence Indic sentiment corpus. Our measurements are of \emph{tokenisers}, and are therefore independent of model precision, hardware, and fine-tuning: they can be reproduced on a laptop in minutes.

\paragraph{The corpus ships with no language labels.} It provides text and sentiment label only. We therefore recover \emph{script} deterministically from Unicode block membership rather than assert language counts we cannot verify. This is the weaker claim, made honestly; it is also the more diagnostic one, since fertility is a property of which codepoints a vocabulary covers, not of linguistic identity. Two conflations follow and we do not paper over them: Devanagari spans Hindi, Marathi and Nepali; Bengali script spans Bengali and Assamese.

\paragraph{It is machine-translated parallel text.} Script counts recover the structure exactly: eleven Indic languages at 10{,}822 sentences \emph{each}, plus 9{,}695 Latin and 1{,}154 Perso-Arabic, summing to precisely 131{,}044. Six scripts share an identical sentence count; Bengali script is exactly $2\times10{,}822$. Corpora do not do this by accident. This is one sentence set rendered into each language.

\paragraph{Two corrections to the record.}
\begin{itemize}
    \item \textbf{Code-mixing is essentially absent}: 345 of 131{,}044 sentences, \textbf{0.3\%}. This dataset is routinely described as containing the code-mixed and Romanised text characteristic of Indic digital communication. It does not. Work using it should claim no code-mixed capability.
    \item \textbf{The class imbalance runs the other way}: 40.1\% positive, 35.1\% negative, 24.8\% neutral. Neutral is the \emph{smallest} class, not the dominant one, so the usual stated justification for weighted F1 on this corpus is wrong. Separately, the Perso-Arabic subset contains \emph{zero} neutral examples.
\end{itemize}

\begin{figure*}[t]
\centering
\includegraphics[width=\textwidth]{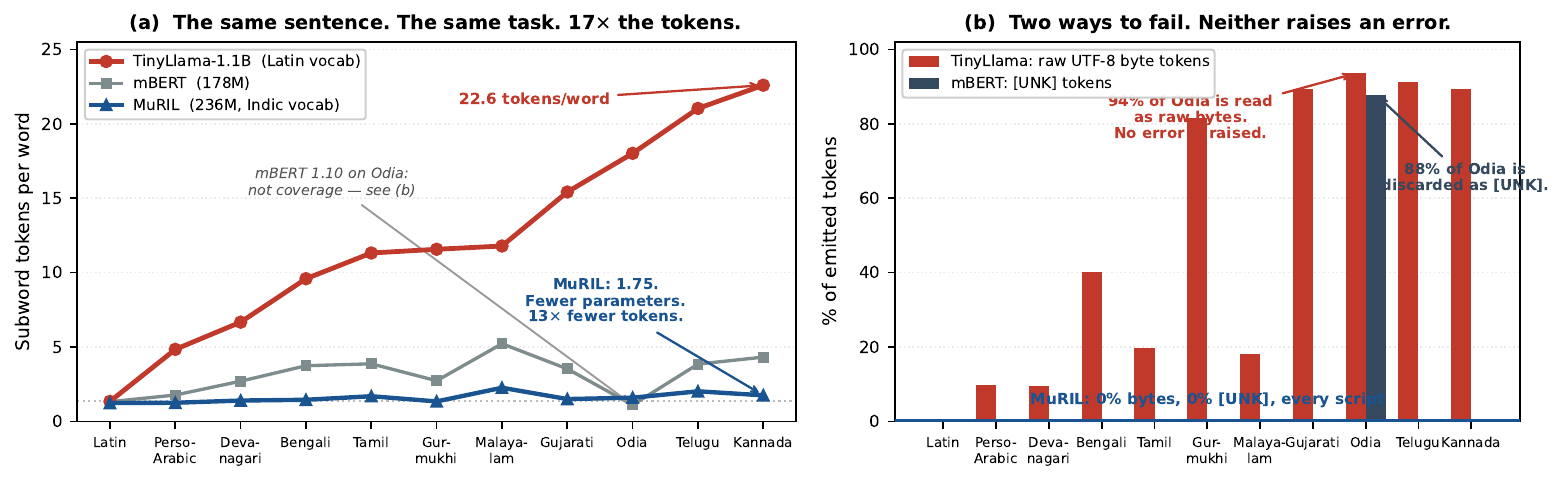}
\caption{\textbf{(a)} Fertility across eleven scripts, sorted. TinyLlama's 32k Latin-trained vocabulary degrades monotonically to 22.60 tokens/word; MuRIL (fewer parameters, but a 197k Indic vocabulary) stays flat. \textbf{(b)} The fraction of emitted tokens the vocabulary cannot represent. TinyLlama substitutes raw UTF-8 bytes (expensive but information-preserving); mBERT destroys Odia as UNK (irrecoverable). Neither raises an error. MuRIL does neither, on any script.}
\label{fig:main}
\end{figure*}

\begin{table*}[t]
\centering\small
\begin{tabular}{@{}lr rrr rrr rrr@{}}
\toprule
& & \multicolumn{3}{c}{\textbf{TinyLlama-1.1B} \emph{(32k vocab)}}
  & \multicolumn{3}{c}{\textbf{mBERT} \emph{(120k vocab)}}
  & \multicolumn{3}{c}{\textbf{MuRIL} \emph{(197k Indic vocab)}} \\
\cmidrule(lr){3-5}\cmidrule(lr){6-8}\cmidrule(lr){9-11}
\textbf{Script} & \textbf{\% corp.}
& \textbf{cov.} & \textbf{fert.} & \textbf{fall.}
& \textbf{cov.} & \textbf{fert.} & \textbf{fall.}
& \textbf{cov.} & \textbf{fert.} & \textbf{fall.} \\
\midrule
Latin        &  7.4 & 50\% &  1.34 &  0.0 & 47\% & 1.32 &  0.0 & 20\% & 1.23 & 0.0 \\
Perso-Arabic &  0.9 & 20\% &  4.84 &  9.8 & 38\% & 1.77 &  0.0 & 47\% & 1.25 & 0.0 \\
Devanagari   & 17.4 & 30\% &  6.67 &  9.6 & 67\% & 2.70 &  0.1 & 72\% & 1.40 & 0.0 \\
Bengali      & 16.5 & 24\% &  9.59 & 40.3 & 75\% & 3.74 &  0.2 & 78\% & 1.45 & 0.0 \\
Tamil        &  8.3 & 26\% & 11.32 & 19.7 & 65\% & 3.87 &  0.0 & 64\% & 1.69 & 0.0 \\
Gurmukhi     &  8.3 &  5\% & 11.57 & \textbf{81.7} & 78\% & 2.73 &  0.0 & 84\% & 1.34 & 0.0 \\
Malayalam    &  8.3 & 28\% & 11.79 & 18.1 & 59\% & 5.22 &  0.0 & 59\% & 2.26 & 0.0 \\
Gujarati     &  8.3 &  1\% & 15.42 & \textbf{89.3} & 78\% & 3.53 &  0.0 & 80\% & 1.50 & 0.0 \\
Odia         &  8.3 &  \textbf{0\%} & \textbf{18.02} & \textbf{93.6} & \textbf{0\%} & \textit{1.10} & \textbf{87.9}$^\dagger$ & 78\% & 1.59 & 0.0 \\
Telugu       &  8.3 &  1\% & 21.04 & \textbf{91.2} & 63\% & 3.85 &  0.0 & 63\% & 2.02 & 0.0 \\
Kannada      &  8.3 &  2\% & \textbf{22.60} & \textbf{89.4} & 80\% & 4.32 &  0.0 & 82\% & 1.75 & 0.0 \\
\midrule
\multicolumn{2}{@{}l}{\textit{fertility, max\,/\,min}} & & \textbf{16.9$\times$} & & & 4.0$\times$ & & & \textbf{1.8$\times$} & \\
\bottomrule
\end{tabular}
\caption{Measured tokenisation, $n{=}400$ sentences per script.
\textbf{cov.}: percentage of the script's assigned Unicode codepoints present in the vocabulary.
\textbf{fert.}: mean subword tokens per whitespace-delimited word.
\textbf{fall.}: percentage of emitted tokens the vocabulary cannot represent: \texttt{<0xXX>} byte tokens for SentencePiece (TinyLlama), UNK for WordPiece (mBERT, MuRIL).
$^\dagger$\,The two fallback mechanisms are \emph{not} equivalent: byte fallback preserves information at ${\approx}3$ tokens per character, whereas UNK destroys it irrecoverably; see Section~5.
Coverage is computed over each full Unicode block and is therefore conservative for Latin, which contains many characters absent from this corpus; it should be read \emph{within} a script, across tokenisers.}
\label{tab:main}
\end{table*}

\section{The Vocabulary Is Empty, Not Inefficient}
\label{sec:vocab}

We report \emph{fertility}: the mean number of subword tokens emitted per whitespace-delimited word~\cite{rust2021tokenizer}.

Llama-2's SentencePiece tokeniser, which TinyLlama inherits, enables byte fallback: a character absent from the vocabulary decomposes into individual \texttt{<0xXX>} tokens. Indic codepoints occupy three UTF-8 bytes. A script with \emph{zero} vocabulary coverage should therefore produce fertility equal to its bytes-per-word, the \textbf{Ceiling} column of Table~\ref{tab:main}. This is a falsifiable prediction, not an impression.

It is confirmed, and not marginally. \textbf{TinyLlama's vocabulary contains 0\% of Odia's codepoints}, and \textbf{93.6\% of the tokens it emits on Odia are raw \texttt{<0xXX>} bytes} (the residual 6.4\% are spaces, digits and ASCII punctuation, which are covered). Its Odia fertility of 18.02 against a byte-fallback ceiling of 18.1 is \textbf{99.6\% of the theoretical maximum}: the two measurements agree, by different routes, that the vocabulary is empty. Telugu (1\% coverage) reaches 94.3\% of its ceiling; Gujarati (1\%) 94.6\%; Kannada (2\%) 90.4\%; Gurmukhi (5\%) 85.1\%. For five of eleven scripts, TinyLlama is not tokenising the text. It is reading bytes.

The gradient is informative. Devanagari, at 30\% coverage, reaches only 40\% of its ceiling; Latin, at 50\%, reaches 21\%. \textbf{Coverage is not rationed by speaker population}: Telugu (1\% coverage) has more speakers than Malayalam (28\%) or Tamil (26\%). The vocabulary tracks not need but web presence, the asymmetry documented by~\cite{joshi2020state}, now embedded, permanently and invisibly, in the tokeniser of a model shipped as multilingual.

\paragraph{Reproducibility.} All quantities in Table~\ref{tab:main} are properties of the \emph{tokeniser alone}. They require no GPU, no fine-tuning and no model weights, and are reproducible in minutes from the public checkpoints \texttt{TinyLlama-1.1B-Chat-v1.0}, \texttt{bert-base-multilingual-cased} and \texttt{google/muril-base-cased}.

\paragraph{MuRIL settles the causal question.} MuRIL~\cite{khanuja2021muril} is an encoder with an Indic vocabulary. On identical text it reads Kannada at \textbf{1.75} tokens/word against TinyLlama's 22.60, a $13\times$ difference, holds every script between 1.23 and 2.26, and emits zero UNK and zero byte tokens throughout. It does so with \emph{a fifth} of TinyLlama's parameters. Even mBERT~\cite{devlin2019bert}, at 178M, has far better Indic coverage (59--80\%) than the 1.1B decoder (0--30\%). The relevant budget is not parameters but \emph{vocabulary}: 32k tokens for TinyLlama against 197k for MuRIL. \textbf{The variable is not scale, architecture, or adaptation method. It is which codepoints made it into the vocabulary.}

This relocates the problem, and the relocation is the point. If the bottleneck were parameters, low-resource communities are priced out of their own fix. If it is the vocabulary, the remedy, vocabulary extension, script-aware tokenisation, or continued in-script pre-training, costs orders of magnitude less than scale and is within reach of a group with no cluster.

\section{The Failure Cannot Be Detected}
\label{sec:silent}

The field's routine diagnostic for tokenisation failure is the UNK rate. It does not work here, and the reason is structural rather than accidental.

Byte fallback exists \emph{precisely so that} a tokeniser can encode any input: a character with no vocabulary entry is decomposed into its UTF-8 bytes rather than replaced by an unknown-token marker. \textbf{A byte-fallback tokeniser therefore cannot emit UNK.} TinyLlama's UNK rate is 0.0\% on Odia, on Kannada, and on every script we tested, and this is not an empirical result but a consequence of the tokeniser's design. Meanwhile 93.6\% of the tokens it emits on Odia carry no lexical content whatsoever.

The standard instrument is thus blind to this failure by construction. No exception is raised, no warning logged, no anomalous count surfaced anywhere in a conventional training or evaluation pipeline. The loss decreases, the model converges, the score is reported. Nothing distinguishes a well-tokenised sequence from one that has silently become seventeen times longer and stripped of lexical structure.

We believe this explains the failure's persistence. A tokeniser that fails \emph{loudly}, as mBERT does on Odia, gets noticed. One that quietly degrades to bytes does not.

\section{Low Fertility Can Mean Destruction}
\label{sec:unk}

mBERT reads Odia at 1.10 tokens per word: better than its 1.32 on Latin, better than MuRIL's 1.59, and the best fertility score anywhere in its column. Ranked on fertility alone, mBERT is the most efficient Odia tokeniser we tested.

It is also destroying the text. Odia is absent from mBERT's vocabulary (0\% coverage) and from its 104 pre-training languages, and \textbf{87.9\% of the tokens it emits on Odia are UNK}. A fertility of ${\approx}1.0$ on an unsupported script is the signature of exactly one UNK per word.

The general form is worth stating plainly, because any efficiency-oriented comparison can fall into it: \textbf{the most catastrophic possible tokenisation failure produces the most attractive possible fertility number.} Total information destruction is indistinguishable, on that metric, from perfect coverage. We would expect this artefact to be present, unnoticed, in other multilingual efficiency comparisons.

\paragraph{The two fallback mechanisms are not equivalent.} We report them in a single column of Table~\ref{tab:main} because both indicate a vocabulary with no entry for the script. Their consequences differ, and the difference is the point. Byte fallback \emph{preserves} the information: a model can in principle learn from a byte sequence, at roughly three tokens per character and with no lexical structure to build upon. UNK \emph{destroys} it: an Odia word collapsed to a single unknown-token marker is unrecoverable, and no downstream capacity restores it. \textbf{TinyLlama is doing something expensive. mBERT is doing something irreversible.}

Odia exhibits both at once: the 1.1B decoder reads 94\% of it as raw bytes, the 178M encoder destroys 88\% of it, and the only model that reads Odia is the one whose vocabulary was built for it.

\section{What 17$\times$ Costs the User}
\label{sec:context}

Fertility is easy to mistake for bookkeeping. It is not. TinyLlama's context window is 2048 tokens, and the \emph{usable} window, in words, is $2048/\text{fertility}$:

\begin{center}\small
\begin{tabular}{@{}lrr@{}}
\toprule
\textbf{Script} & \textbf{Words of context} & \textbf{vs.\ Latin} \\
\midrule
Latin      & 1{,}528 & 1.00$\times$ \\
Devanagari &    307 & 0.20$\times$ \\
Bengali    &    214 & 0.14$\times$ \\
Malayalam  &    174 & 0.11$\times$ \\
Odia       &    114 & 0.07$\times$ \\
Telugu     &     97 & 0.06$\times$ \\
Kannada    &  \textbf{91} & \textbf{0.06$\times$} \\
\bottomrule
\end{tabular}
\end{center}

\medskip
\noindent
\textbf{The same model, on the same hardware, gives an English user 1{,}528 words of context and a Kannada user 91.} Nothing in the model card, the parameter count, or the memory benchmark discloses this. It is not a limitation the Kannada user can budget around, and it is not one they are told about.

\paragraph{The cost compounds under load.} TinyLlama's KV cache requires ${\approx}22$\,KB per token (22 layers, 4 grouped-query key-value heads, head dimension 64, FP16). A 30-word sentence is ${\approx}75$ tokens in Latin script and ${\approx}713$ in Kannada. At batch size 1 this is a 14\,MB difference and weight memory dominates, so a memory benchmark run on short Latin inputs will show quantisation working exactly as advertised. At batch size 32, the same gap is 52\,MB against 490\,MB. \textbf{The reported efficiency of these systems is thus most flattering precisely where they are least used: single-sentence, English, short-input benchmarks.} We flag this as an analytical consequence of the measured fertility; establishing it empirically across serving regimes is immediate future work.

\section{Discussion and Limitations}

\paragraph{What we would ask of the field.} Report efficiency per script, with the script distribution of the evaluation set. Report fertility with an UNK/byte-fallback rate; without it, the metric cannot distinguish coverage from destruction. And when an SLM is proposed for a low-resource language, \textbf{check the vocabulary before the parameter count}. It requires no GPU and five minutes.

\paragraph{Limitations.} (i) \textbf{Script is not language}, and the two conflations noted in \S\ref{sec:setup} are real. Our claims are about scripts. (ii) \textbf{We measure the mechanism, not yet the downstream harm.} We establish that the tokenisation of five Indic scripts collapses to raw bytes, and that the usable context window shrinks $17\times$; we do not here report the task-accuracy cost. That measurement is in progress and is the natural next step, but it is a separate claim and we do not want to imply we have made it. (iii) \textbf{The corpus is machine-translated} parallel text, adequate for tokenisation, which is a property of the script rather than of the naturalness of the sentence, but inadequate for claims about real-world usage, which we do not make.

\paragraph{Conclusion.} Small Language Models are offered to low-resource communities as the affordable path to multilingual AI. For five of the eleven Indic scripts examined here, the model on offer contains no vocabulary for the script at all, reads it as raw bytes, reports no error while doing so, and hands its users a context window a sixteenth the size of an English speaker's. The mechanism is not scale. It is a vocabulary, it is measurable on a laptop in five minutes, and it is cheap to fix. That is what we would like this community to take away.

\section*{Ethical Statement}
A tokeniser that encodes Kannada at seventeen times the cost of English imposes a measurable tax, in context, latency and money, on the speakers least able to bear it~\cite{ahia2023cost,petrov2023tokenizers}. That this tax is invisible to every standard metric, and that the affected communities are given no means to detect it, is in our view the more serious finding. We also note that the corpus studied here presents machine-translated text as multilingual data, and that its code-mixing content is 0.3\% rather than substantial: such resources are readily mischaracterised, including by us in earlier work.

\section*{Acknowledgements}
The author thanks the anonymous reviewers of the GlobalSouthAI workshop for their careful and constructive feedback, and the workshop organisers for providing a venue for this work.


\begin{thebibliography}{}

\bibitem[Ahia et al., 2023]{ahia2023cost}
Orevaoghene Ahia, Sachin Kumar, Hila Gonen, Jungo Kasai, David~R. Mortensen, Noah~A. Smith, and Yulia Tsvetkov.
\newblock Do all languages cost the same? Tokenization in the era of commercial language models.
\newblock In {\em Proc.\ EMNLP}, 2023.

\bibitem[Devlin et al., 2019]{devlin2019bert}
Jacob Devlin, Ming-Wei Chang, Kenton Lee, and Kristina Toutanova.
\newblock {BERT}: Pre-training of deep bidirectional transformers for language understanding.
\newblock In {\em Proc.\ NAACL-HLT}, pages 4171--4186, 2019.

\bibitem[Joshi et al., 2020]{joshi2020state}
Pratik Joshi, Sebastin Santy, Amar Budhiraja, Kalika Bali, and Monojit Choudhury.
\newblock The state and fate of linguistic diversity and inclusion in the {NLP} world.
\newblock In {\em Proc.\ ACL}, 2020.

\bibitem[Khanuja et al., 2021]{khanuja2021muril}
Simran Khanuja, Diksha Bansal, Sarvesh Mehtani, Savya Khosla, Atreyee Dey, Balaji Gopalan, Dilip~Kumar Margam, Pooja Aggarwal, Rajiv~Teja Nagipogu, Shachi Dave, Shruti Gupta, Subhash Chandra~Bose Gali, Vish Subramanian, and Partha Talukdar.
\newblock {MuRIL}: Multilingual representations for {I}ndian languages.
\newblock {\em arXiv:2103.10730}, 2021.

\bibitem[Petrov et al., 2023]{petrov2023tokenizers}
Aleksandar Petrov, Emanuele La~Malfa, Philip H.~S. Torr, and Adel Bibi.
\newblock Language model tokenizers introduce unfairness between languages.
\newblock In {\em Proc.\ NeurIPS}, 2023.

\bibitem[Rust et al., 2021]{rust2021tokenizer}
Phillip Rust, Jonas Pfeiffer, Ivan Vuli\'{c}, Sebastian Ruder, and Iryna Gurevych.
\newblock How good is your tokenizer? On the monolingual performance of multilingual language models.
\newblock In {\em Proc.\ ACL}, 2021.

\bibitem[Schwartz et al., 2020]{schwartz2020green}
Roy Schwartz, Jesse Dodge, Noah~A. Smith, and Oren Etzioni.
\newblock Green {AI}.
\newblock {\em Communications of the ACM}, 63(12):54--63, 2020.

\bibitem[Strubell et al., 2019]{strubell2019energy}
Emma Strubell, Ananya Ganesh, and Andrew McCallum.
\newblock Energy and policy considerations for deep learning in {NLP}.
\newblock In {\em Proc.\ ACL}, 2019.

\bibitem[Zhang et al., 2024]{zhang2024tinyllama}
Peiyuan Zhang, Guangtao Zeng, Tianduo Wang, and Wei Lu.
\newblock {TinyLlama}: An open-source small language model.
\newblock {\em arXiv:2401.02385}, 2024.

\end{thebibliography}
\end{document}